\pdfoutput=1

\documentclass[11pt]{article}

\usepackage[]{acl}

\usepackage{times}
\usepackage{latexsym}
\usepackage{lipsum} 
\usepackage{graphicx}
\usepackage{graphicx}
\usepackage{graphicx}
\usepackage{subfigure}
\usepackage{amsmath}
\usepackage{amssymb}

\usepackage{latexsym}
\usepackage{graphicx}

\usepackage{graphicx}
\usepackage{amsmath}
\usepackage{amsthm}
\usepackage{booktabs}
\usepackage{algorithm}
\usepackage{algorithmic}
\usepackage{multirow}
\usepackage{amsmath}
\usepackage{amssymb}
\usepackage{graphicx}
\usepackage{graphicx}
\usepackage{subfigure}
\usepackage{times}
\usepackage{latexsym}
\usepackage{algorithmic}
\usepackage{graphicx}
\usepackage{graphicx}
\usepackage{subfigure}
\usepackage{amsmath}
\usepackage{amssymb}
\usepackage{algorithm}
\usepackage{algorithmic}

\usepackage{graphicx}
\usepackage{amsmath}
\usepackage{amsthm}
\usepackage{booktabs}
\usepackage{algorithm}
\usepackage{algorithmic}
\usepackage{multirow}
\usepackage{amsmath}
\usepackage{amssymb}
\usepackage{graphicx}
\usepackage{graphicx}
\usepackage{subfigure}

\usepackage{graphicx}
\usepackage{graphicx}
\usepackage{subfigure}
\usepackage{amsmath}
\usepackage{amssymb}
\usepackage{algorithm}
\usepackage{algorithmic}
\usepackage{latexsym}
\usepackage{lipsum}

\usepackage{graphicx}
\usepackage{amsmath}
\usepackage{amsthm}
\usepackage{booktabs}
\usepackage{algorithm}
\usepackage{algorithmic}
\usepackage{multirow}
\usepackage{amsmath}
\usepackage{amssymb}
\usepackage{graphicx}
\usepackage{graphicx}
\usepackage{subfigure}
\usepackage{lipsum}
\usepackage{wasysym}
\usepackage{bm}
\usepackage{algorithm}

\usepackage{microtype}
\usepackage{tabularx}

\usepackage{amsmath}

\usepackage[T1]{fontenc}

\usepackage[utf8]{inputenc}

\usepackage{microtype}

\usepackage{inconsolata}

\usepackage{graphicx}

\title{Event Argument Extraction with Enriched Prompts}

\author{Chen Liang \\
 Beijing Jiaotong University, Beijing, China \\ 
	\tt	\{nlp\_liangchen\}@bjtu.edu.cn
}

\begin{document}
\maketitle
\begin{abstract}
This work aims to delve deeper into prompt-based event argument extraction (EAE) models.
We explore the impact of incorporating various types of information into the prompt on model performance, including trigger, other role arguments for the same event, and role arguments across multiple events within the same document.
Further, we provide the best possible performance that the prompt-based EAE model can attain and demonstrate such models can be further optimized from the perspective of the training objective.
Experiments are carried out on three small language models and two large language models in RAMS.
The code is publicly available at: \url{https://github.com/cs-liangchen-work/EAEPrompt/tree/main}.
\end{abstract}

\section{Introduction}

\begin{figure*}[ht]
\centering
\includegraphics[scale=0.46]{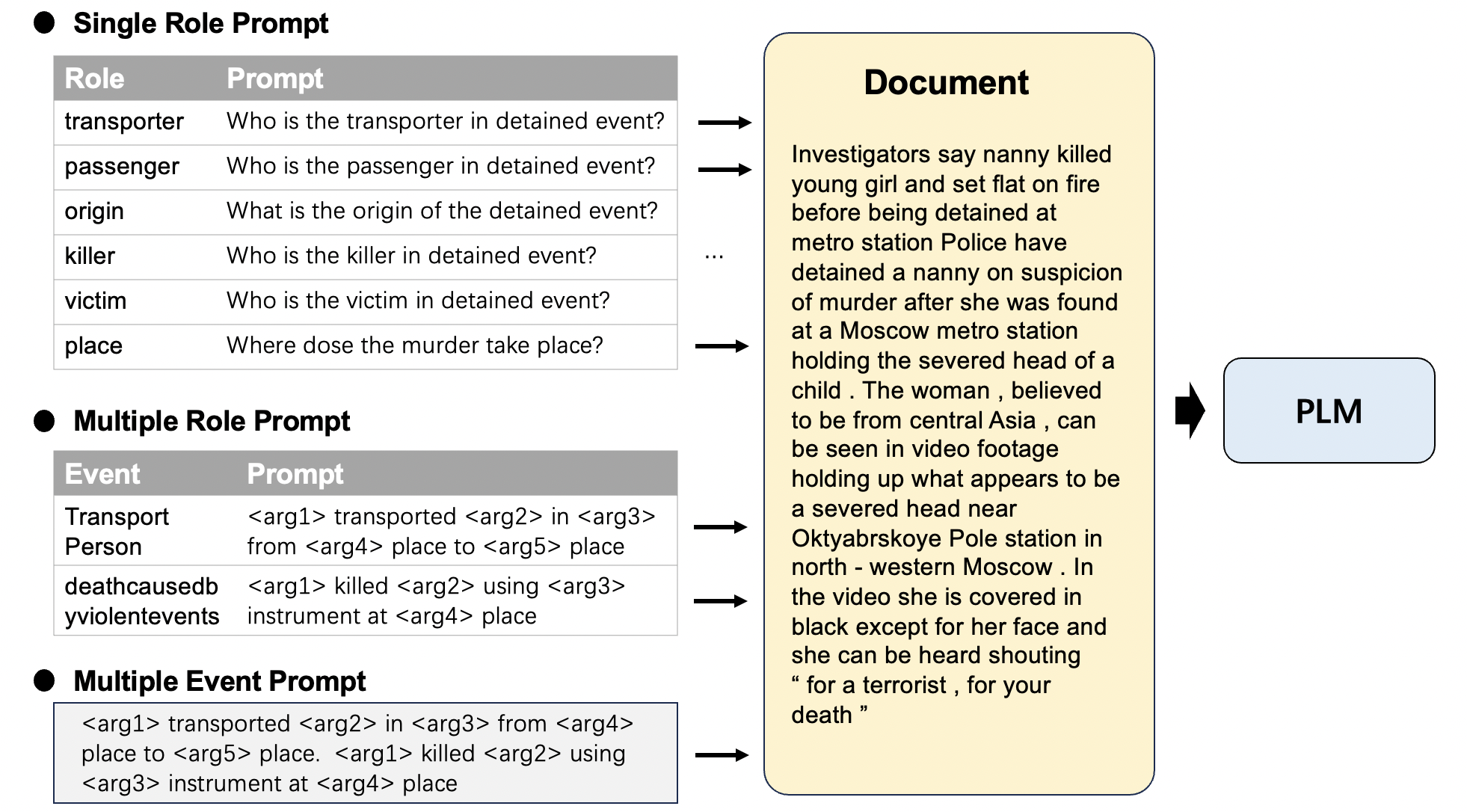}
\caption{The overview of prompt-based EAE model.}
\label{fig_model}
\end{figure*}

Event argument extraction (EAE) aims at discovering role arguments related to the event trigger \cite{li-etal-2021-document, xu-etal-2022-two, zhou-mao-2022-document, liu-etal-2023-document}.
In this task, an event triggered by a specific trigger can contain multiple arguments, and a given document may include several events.
Consider the document present in Figure \ref{fig_model}, it has two event \textit{transport.person} and \textit{ death.caused.by.violent.events}.
The former requires identifying \textit{transporter}, \textit{passenger}, and \textit{origin},  while the latter focuses on locating \textit{killer}, \textit{victim}, and \textit{place}.

Currently, the prompt-based EAE models have demonstrated state-of-the-art effectiveness, benefiting from the understanding of role semantics and introducing enriched information to enhance reasoning process.
Typical efforts can be classified into three groups based on the information provided in the prompt:
single role prompt model \cite{liu-etal-2020-event, du-cardie-2020-event}, 
multiple roles prompt model \cite{li-etal-2021-document,wei-etal-2021-trigger, ma-etal-2022-prompt},
and multiple events prompt model \cite{liu-etal-2024-beyond-single}.
As the prompt-based EAE model continues to be optimized and it's performance improves, an important question emerges: What is the performance ceiling for this model type?

In this paper, we investigate the best possible performance that the prompt-based EAE model can attain.
For the multiple roles prompt model, we think all other role arguments for the same event acting as clues can offer the utmost support in locating the target argument.
Similarly, the multiple events prompt model achieves the maximum performance when all other role arguments across multiple events within the same document are used as clues.
Moreover, we propose a loss regularization technique to strengthen the prompt-based model, suggesting that such models can be further optimized from the perspective of the training objective.

By conducting extensive experiments on BERT \cite{devlin-etal-2019-bert}, BART \cite{lewis2019bart}, Roberta \cite{liu2019roberta}, Llama-3 \cite{dubey2024llama}, and GPT-4 \cite{achiam2023gpt}, we conclude that: 
$i)$ The performance of the current prompt-based EAE model can still be improved, for example, by using other arguments as clues or modifying the loss function.
$ii)$ The effect of intra-event information is more substantial than that of inter-event information.
$iii)$ The current model cannot fully comprehend the additional information in the prompt.

\section{Related Work}
Event argument extraction (EAE) \cite{li-etal-2021-document, xu-etal-2022-two, zhou-mao-2022-document, liu-etal-2023-document} aims at locating arguments in texts with event types.
Recently, prompt learning \cite{ma-etal-2022-prompt} has exhibited remarkable effectiveness in the EAE task.
Prompt-based EAE model is first proposed by \citet{liu-etal-2020-event, du-cardie-2020-event}, which constructs questions for each role and jointly encodes them with the document to identify arguments.
Later, more enriched information is injected into the prompt to boost the model. \citet{ma-etal-2022-prompt} propose multi-role prompts to capture argument correlations. \citet{liu-etal-2024-beyond-single} present a multi-event prompt technology to model the interactions among multiple events.
Moreover, prompt-based models, by leveraging their understanding of the role semantics, can incorporate out-of-domain datasets to augment the training data \cite{liu-etal-2021-machine,liu2022document,chen2023ntda}.
In the era of large language model (LLM), researchers \cite{zhou-etal-2024-llms,fu-etal-2024-tise} utilize LLM's instruction (prompt) following and in-context learning abilities 
achieve superior performance with several demonstrations and even surpass supervised models. 
In this paper, we summarize prompt-based EAE models with varying degrees of information density and investigate their performance limitations, aiming to facilitate future research.

\section{Method}
This section first introduces three EAE models with gradually increased information in the prompt ($\S$\ref{sec:m1}, $\S$\ref{sec:m2}, $\S$\ref{sec:m3}). Finally, we present a loss regularization method to enhance the prompt-based model ($\S$\ref{sec:m4}).


\subsection{Single Role Prompt Model}
\label{sec:m1}
\textbf{Information:} \textit{role and trigger.}

Given the event document $D$ and the target role $r$ to be identified, the input sequence in the prompt learning paradigm \cite{brown2020language} is as follows:
\begin{align}
      Q(r) \  [SEP] \  D
\end{align}
where $Q(\cdot)$ is the natural question generation strategy for $r$ \cite{liu-etal-2020-event,du-cardie-2020-event}.

Different from other information extraction tasks, there is an interaction between the trigger and the role in EAE.
In NER task \cite{li-etal-2020-unified}, a role label may correspond to multiple arguments. However, in EAE, due to the constraints of the trigger, argument extraction must account for both the trigger and the role.
Trigger information can be introduced in the following ways:
\begin{itemize}
    \item{use BERT's \cite{devlin-etal-2019-bert} segment embedding to distinguish the trigger (set to 1) from other tokens (set to 0) \cite{ebner-etal-2020-multi}.} 
    \item{define text markers <t> and </t> to indicate the trigger \cite{li-etal-2021-document,ma-etal-2022-prompt}.}
    \item{included in the prompt \cite{liu-etal-2020-event,du-cardie-2020-event}. }
\end{itemize}

\subsection{Multiple Roles Prompt Model}
\label{sec:m2}
\textbf{Information:} \textit{other role argument for the same event.}

Considering that roles within the same event are interrelated, rather than being independent, we can leverage related roles of the target as clues to help its extraction\cite{wei-etal-2021-trigger, ma-etal-2022-prompt}, as follows:
\begin{align}
      T(r_1, r_2, r_3 ...) \  [SEP] \  D
\end{align}
where $r_i$ is roles within the same event, $T(\cdot)$ is templates for modeling the connections among roles.

\paragraph{Upper Bound}
The upper bound of the multiple roles prompt model's performance is when the other arguments within the same event are available and integrated into the template as clues while extracting one target role argument.

\subsection{Multiple Events Prompt Model}
\label{sec:m3}
\textbf{Information:} \textit{role arguments across multiple events within the same document.}

Previous works \cite{liang-etal-2022-raat, liu-etal-2024-beyond-single} have confirmed that cross-event information can benefit EAE.
We organize the input into the following sequence to capture the beneficial event correlations:
\begin{align}
      T_{m}(T_{1}(\cdot), T_{2}(\cdot), T_{3}(\cdot) ...) \  [SEP] \  D
\end{align}
where $T_{m}(\cdot)$ is used to concatenate different event templates.
\paragraph{Upper Bound}
The model's upper bound is achieved when the other arguments in the document are known (maybe have several events) and are filled into the template as clues during the extraction of the target role arguments.

\subsection{Enhanced EAE Model}
\label{sec:m4}
Inspired by \citet{chen2023ntda} employing the region loss function to achieve self-denoising, we believe that all prompt-based EAE models can benefit from this loss and have conducted extensive experiments to validate this conclusion.
The region loss function we used as follows:
\begin{align}
loss_{dice} = 1 - \frac{2\sum\limits_{i=1}^{N}p_iq_i}{\sum\limits_{i=1}^{N}p_i^2 + \sum\limits_{i=1}^{N}q_i^2}
\end{align}
where $p_i$ and $q_i$ is the $i$-th value for gold and predicted label, $N$ denotes the length of the document.

\section{Experimental Setups}
\paragraph{Dataset}
We evaluated the effectiveness of our method on RAMS \cite{ebner-etal-2020-multi}, which is a widely used document-level benchmark in the EAE task. RAMS contains 139 event types, 63 role types, and 7,239 documents.
\paragraph{Baseline}
QAEE \cite{du-cardie-2020-event} and DocMRC \cite{liu-etal-2021-machine}, which are single-role prompt models. 
PAIE \cite{ma-etal-2022-prompt}, which is multiple roles prompt model. 
DEEIA \cite{liu-etal-2024-beyond-single}, which is multiple events prompt model.

\section{Experimental Results}

\setlength{\tabcolsep}{2.8mm}{
\begin{table*}
\centering
\begin{tabular}{llcccccc}
\toprule
\multirow{2}{*}{\textbf{Model}} &  \multirow{2}{*}{\textbf{PLM}} & \multicolumn{3}{c}{\textbf{RAMS}} & \multicolumn{3}{c}{\textbf{RAMS} w dice} \\
 \cmidrule(lr){3-5} \cmidrule(lr){6-8} 
 & & P & R & F1 & P & R & F1  \\ 
\hline
QAEE \cite{du-cardie-2020-event} & BERT-b & 42.4 & 44.9 & 43.6 & - & - & - \\
DocMRC \cite{liu-etal-2021-machine} & BERT-b & 43.4 & 48.3 & 45.7 & - & - & -  \\
\multirow{2}{*}{PAIE \cite{ma-etal-2022-prompt}}
 & BART-b & - & - & 49.5 & - & - & - \\
 & BART-l & - & - & 52.2 & - & - & - \\
DEEIA \cite{liu-etal-2024-beyond-single} & Roberta-l & - & - & 53.4 & - & - & -  \\
\hline
\multirow{3}{*}{R-Prompt \cite{chen2023ntda}}
 & BERT-b & 43.9 & 42.1 & 42.5 & 43.2 & 42.9 & 43.1  \\
 & BART-b & 43.1 & 47.4 & 45.1 & 44.0 & 47.6 & 45.7 \\
 & Roberta-l & 47.3 & 49.2 & 48.2 & 45.0 & 53.3 & 48.8  \\
\cline{2-8} 
 \multirow{3}{*}{mRole-Prompt}
 & BERT-b & 43.8 & 49.1 & 46.3 & 45.8 & 47.8 & 46.8  \\
 & BART-b & 45.2 & 52.7 & 48.7 & 45.4 & 53.6 & 49.1  \\
 & Roberta-l & 46.7 & 56.3 & 51.0 & 47.1 & 56.4 & 51.3 \\
 \cline{2-8} 
 \multirow{3}{*}{mR-Prompt (ceiling)}
 & BERT-b & 44.2 & 52.2 & 47.9 & 46.2 & 51.5 & 48.7  \\
 & BART-b & 47.6 & 53.6 & 50.4 & 49.8 & 52.8 & 51.3 \\
 & Roberta-l & 50.4 & 61.3 & 55.3 & 53.6 & 58.3 & 55.8 \\
\cline{2-8} 
 \multirow{3}{*}{mEvent-Prompt }
 & BERT-b & 40.5 & 53.2 & 46.0 &  43.1 & 50.0 & 46.3 \\
 & BART-b & 43.8 & 53.5 & 48.2 & 45.3 & 52.6 & 48.8  \\
 & Roberta-l & 45.9 & 56.4 & 50.6 & 49.9 & 52.9 & 51.4 \\
 \cline{2-8} 
  \multirow{3}{*}{mE-Prompt (ceiling)}
 & BERT-b & 43.1 & 54.8 & 48.2 &  44.9 & 52.5 & 48.4  \\
 & BART-b & 46.5 & 57.1 & 51.2 &  46.4 & 58.6 & 51.8 \\
 & Roberta-l & 55.3 & 55.6 & 55.4 & 48.3 & 67.9 & 56.4 \\
\hline
 \cline{2-8} 
 Prompt testing & BERT-b & 95.2 & 96.3 & 95.7 & - & - & -   \\
\toprule
\end{tabular}
\caption{Experimental results of different prompt-based EAE model.}
\label{tab:main_1}
\end{table*}
}

We present the main results in Table 1.
R-Prompt model is the single role prompt model as described in $\S$\ref{sec:m1}.
mRole-Prompt model and mR-Prompt (ceiling) are the multiple roles prompt model and its performance upper bound model, as shown in $\S$\ref{sec:m2}.
mEvent-Prompt model and mE-Prompt (ceiling) are the multiple events prompt model and its ceiling performance model, as described in $\S$\ref{sec:m3}.
'w dice' means training model with the dice loss function.
The experiments are carried out in BERT \cite{devlin-etal-2019-bert}, BART \cite{lewis2019bart}, and Roberta \cite{liu2019roberta}.

By comparing the results of the R-prompt model and the mRole-Prompt model, we find that considering the interactions between roles can improve model performance, resulting in a
3.4\% absolute improvement in F1 on average.
The mEvent-Prompt model integrates inter-event role dependencies, but its performance does not outperform the mRole-Prompt model (performance drops 0.4\% in F1).
However, the performance of mE-Prompt (ceiling) surpasses mR-Prompt (ceiling), which implies that while introducing inter-event relationships is beneficial, effective prompt design is needed to fully exploit this information. 
Moreover, the performance improvement of the mR-Prompt (ceiling) model over the R-prompt model is significantly greater than the improvement of the mE-Prompt (ceiling) model over the mR-Prompt (ceiling) model, showing +6.0\% improvement and +0.4\% gains in F1, respectively.
Therefore, we can conclude that \textbf{the effect of intra-event information is more substantial than that of inter-event information}.

We also conduct experiments on two large language models, and the results are presented in Table 2. Due to resource constraints, we randomly selected only 50 samples from the test set for evaluation. To our surprise, the performance of the prompt-based EAE model on LLM significantly diverges from what we predicted.
One reason is hallucination issues in LLM.

Table 1 presents several of the best-performing EAE baseline models, and we can observe that their performance still shows a gap compared to the mE-Prompt (ceiling) model and mR-Prompt (ceiling) model. What strategies can be employed to further improve performance?
From the results in Table 1, we can see that leveraging the complex dependencies between roles is important for improving model performance. 
By comparing the performance of models with the same prompt across different language models (BERT, BART, and Roberta), it is evident that the underlying model's capabilities also play a crucial role.
By comparing the results of models trained with and without dice loss, we observe that 
all prompt-based EAE models can benefit from this loss. 
Therefore, we can conclude that \textbf{the performance of the current prompt-based EAE model can still be improved, for example, by using other arguments as clues or modifying the loss function.}

'Prompt testing' is an experiment in which gold arguments are appended to the prompt to assess the model's ability to understand the prompt, yielding 95.7\% F1. We can conclude that \textbf{the current model cannot fully comprehend the additional information in the prompt}.

\setlength{\tabcolsep}{3.5pt}
\begin{table}
\centering
\begin{tabularx}{\linewidth}{Xccc}
\toprule
\textbf{Setting} & \textbf{P}  & \textbf{R} & \textbf{F1} \\ \midrule 
Llama-3 (70B) &  &  &    \\
\ \ \ \ \ \ Role-prompt & 40.3 & 58.2 & 47.7  \\
\ \ \ \ \ \ mRole-prompt & 46.7 & 45.9 & 46.3  \\
\ \ \ \ \ \ mR-prompt (ceiling) & 39.2 & 54.9 & 45.7   \\
\hline
GPT-4 &  &  &    \\
\ \ \ \ \ \ Role-prompt & 50.4 & 56.6 & 53.3  \\
\ \ \ \ \ \ mRole-prompt & 53.1 & 55.7 & 54.4  \\
\ \ \ \ \ \ mR-prompt (ceiling) &  49.6 & 55.7 & 52.5  \\
\bottomrule
\end{tabularx}
\caption{Results on large language model.}
\label{tab:ablation}
\end{table}

\section{Conclusion}
In this paper, we explore the prompt-based event argument extraction (EAE) model in depth, and provide three conclusions.  We do extensive experiments to verify our conclusions. We hope that our findings will contribute to future research.

\bibliography{arxiv}

\end{document}